\documentclass[sigconf,natbib=true]{acmart}

\usepackage{booktabs} 
\usepackage{multirow}
\usepackage{subfigure}
\usepackage{verbatim}
\usepackage{CJKutf8}
\usepackage{flushend}

\usepackage{makecell}
\usepackage[utf8]{inputenc}

\copyrightyear{2022}
\acmYear{2022}
\setcopyright{acmcopyright}
\acmPrice{15.00}

\begin{document}

\title{Semi-FairVAE: Semi-supervised Fair Representation Learning with Adversarial Variational Autoencoder}

\fancyhead{}


\author{Chuhan Wu$^1$, Fangzhao Wu$^2$, Tao Qi$^1$, Yongfeng Huang$^1$}

\affiliation{%
  \institution{$^1$Department of Electronic Engineering, Tsinghua University, Beijing 100084 \\ $^2$Microsoft Research Asia, Beijing 100080, China}
} 
\email{{wuchuhan15,wufangzhao,taoqi.qt}@gmail.com,yfhuang@tsinghua.edu.cn}






\begin{abstract}

Adversarial learning is a widely used technique in fair representation learning to remove the biases on sensitive attributes from data representations.
It usually requires incorporating the sensitive attribute labels as prediction targets.
However, in many scenarios the sensitive attribute labels of many samples can be unknown, and it is difficult to train a strong discriminator based on the scarce data with observed attribute labels, which may lead to generate unfair representations.
In this paper, we propose a semi-supervised fair representation learning approach based on an adversarial variational autoencoder, which can reduce the dependency of adversarial fair models on data with labeled sensitive attributes.
More specifically, we use a bias-aware model to capture inherent bias information on sensitive attributes by accurately predicting sensitive attributes from input data, and use a bias-free model to learn debiased fair representations by using adversarial learning to remove bias information from them.
The hidden representations learned by the two models are regularized to be orthogonal.
In addition, the soft labels predicted by the two models are further integrated into a semi-supervised variational autoencoder to reconstruct the input data, and we apply an additional entropy regularization to encourage the attribute labels inferred from the bias-free model to be high-entropy.
In this way, the bias-aware model can better capture  attribute information while the bias-free model is less discriminative on sensitive attributes if the input data is well reconstructed.
Extensive experiments on two datasets for different tasks validate that our approach can achieve good representation learning fairness under limited data with sensitive attribute labels.

\end{abstract}

%
%


\maketitle

\section{Introduction}

Representation learning models usually aim to build representations of input data by mining its inherent characteristics~\cite{bengio2013representation}.
However, the raw data (e.g., user-generated web data) may encode biases related to some sensitive attributes such as demographics~\cite{zhang2018mitigating}.
The models learned on such data may also inherit these biases and generate biased representations~\cite{madras2018learning}.
For example, researchers have found that in many canonical word embeddings such as Word2vec~\cite{mikolov2013distributed} and GloVe~\cite{pennington2014glove}, the word ``doctor'' has a closer relation to ``male'' while ``nurse'' has a smaller distance to ``female''~\cite{zhao2018learning}.
Such ``stereotypes'' of models may lead to biased decisions that are unfair to groups with certain sensitive attributes~\cite{barocas2017fairness}.
Thus, fair representation learning, which aims to learn representations that are free from the influence of targeted sensitive attributes, is important to build responsible intelligent web systems and make fair  automatic decisions~\cite{mehrabi2021survey}.

Adversarial learning is a widely used fair representation learning technique that aims to remove the information related to sensitive attributes from hidden representations~\cite{lowd2005adversarial}.
It typically uses a discriminator to infer sensitive attributes from the representations learned by a model, and propagates negative gradients of attribute prediction loss to the model to help generate sensitive attribute-agnostic representations~\cite{zhang2018mitigating}.
However, in many real-world scenarios, the sensitive attribute labels of many samples can be missing.
For example, most users on the YouTube platform do not provide their gender information~\cite{filippova2012user}.
In addition, it is very expensive and even impractical to collect or manually annotate sufficient data with sensitive attribute labels~\cite{hu2007demographic}.
The discriminator in adversarial learning cannot be well-tuned if samples with labeled attributes are scarce, and therefore the bias information related to sensitive attributes encoded in the deep representations cannot be effectively eliminated to achieve good fairness.

Semi-supervised learning is an effective way to enhance model performance  by mining useful information from redundant unlabeled data~\cite{van2020survey}.
There are a few attempts to incorporate semi-supervised learning techniques into fair machine learning models~\cite{noroozi2019leveraging}.
For example, Noroozi et al.~\shortcite{noroozi2019leveraging} proposed to use the model to annotate pseudo labels for unlabeled data, and then add samples with high confidence to the training set.
They applied fairness regularization losses to both real and pseudo labeled samples.
Zhang et al.~\shortcite{zhang2020fairness} proposed an iterative method that first combines the raw dataset and the pseudo labeled dataset, then re-samples the data to ensure that the number of samples with different sensitive attributes are equal, and finally trains the model on the re-sampled dataset to further generate the pseudo labeled dataset.
However, these methods mainly aim to handle the scarcity of labeled data in downstream tasks, while they assume that the sensitive attributes of all samples are observed.
It is still very challenging to learn fair models if only a small subset of samples are associated with observed sensitive attribute labels.

In this paper, we proposed a semi-supervised fair representation learning method named \textit{Semi-FairVAE}, which can effectively reduce the dependency of adversarial fair representation learning methods on labeled samples with sensitive attributes via a semi-supervised adversarial variational autoencoder.
Different from the standard semi-supervised variational autoencoder (Semi-VAE) that incorporates the predicted label as the latent variable for input construction, in fair adversarial learning methods the predicted attributes are enforced to be random and it is not suitable to directly take them as latent variables.
Motivated by the decomposed adversarial learning method proposed in~\cite{wu2021fairness}, we use a bias-aware model to capture the bias information  related to sensitive attributes in the input data and use a bias-free model to learn fair representations with bias information eliminated.
The bias-aware model is used to accurately infer the sensitive attributes by training in an attribute prediction task, while the bias-free model is used to learn bias-independent feature representations by applying an adversarial learning module to its output, where the hidden representations learned by the two models are regularized to be orthogonal to further remove bias information from the bias-free model.
In addition, the soft attribute labels predicted by the two models and the feature learned by a VAE encoder are further incorporated as the latent variables in a semi-supervised variational autoencoder to reconstruct the input data.
Note that on samples with labeled attributes, the attribute labels predicted by the bias-aware and bias-free models are respectively replaced by the real labels and soft labels with uniform class probabilities, which aim to provide supervision information on the real and adversarial label distributions to help the input reconstruction in generative semi-supervised learning.
To further encourage the bias-free model to be attribute insensitive, on samples without  attribute labels the attribute predictions from the bias-free model are regularized to have higher entropy so that less sensitive attribution information can be encoded in the bias-free feature representations. 
Extensive experiments on two datasets for different tasks  validate that our proposed \textit{Semi-FairVAE} approach can achieve both good accuracy and fairness under limited data with observed sensitive attribute labels.

The contributions of this paper are listed as follows:
\begin{itemize}
    \item To our best knowledge, this is the first semi-supervised fair representation learning method that aims to handle the scarcity of samples with labeled sensitive attributes.
    \item We propose a semi-supervised adversarial variational autoencoder for semi-supervised fair representation learning, which can effectively exploit data without observed sensitive attribute labels.
    \item We conduct extensive experiments on two datasets for different tasks to verify the effectiveness of our approach in semi-supervised fair representation learning.
\end{itemize}

\section{Related Work}\label{sec:RelatedWork}

\subsection{Fair Representation Learning}

Learning fair representations from the data that encodes biases related to certain sensitive attributes is a widely studied problem in machine learning~\cite{mcnamara2019costs}.
Some early studies explore improving the fairness of representations by adjusting the raw dataset into a fair one for model training~\cite{calders2009building,kamiran2012data,calmon2017optimized}.
For example, Calders et al.~\shortcite{calders2009building} proposed two methods to balance the dataset over different sensitive attributes.
The first one is massaging, which changes the labels of  samples with less confident model predictions.
The second one is reweighting, which assigns samples in model training different weights according to  the proportion of different sensitive attributes over different classes.
Kamiran et al.~\shortcite{kamiran2012data} proposed a preferential sampling method that duplicates or removes the samples that are close to the decision boundary based on the attribute distributions.
These dataset modification methods are usually compatible with different tasks and methods.
However, these methods mainly focus on eliminating the effects of biases encoded by training data, and they cannot handle the potential biases and unfairness brought by the representation learning algorithms.

Another widely used fair representation learning paradigm is adding fairness constraints to representations to regularize  the model~\cite{kamishima2012fairness,zafar2017fairness,donini2018empirical}.
For example, Zemel et al.~\shortcite{zemel2013learning} proposed to use autoencoder to learn hidden representations of input data.
They added a statistical parity regularization to reduce the discrimination of hidden representations on sensitive attributes.
Yao et al.~\shortcite{yao2017beyond}  derived four different fairness metrics from the predicted and real ratings of users with different sensitive attributes, and they compared regularizing the collaborative filtering models with one of these metrics.
These regularization based methods can control the tradeoff between accuracy and fairness by choosing different regularization intensities.
However, the fairness constraints are often difficult to achieve and may even contradict the objectives of the main prediction task, and thus the model optimization may not be effective.
In addition, in many real-world applications such as click-through rate prediction and news recommendation, it is difficult to design proper fairness constraints for model training~\cite{wu2021fairness}.

In recent years, adversarial learning becomes a new fashion in fair representation learning~\cite{zhang2018mitigating,madras2018learning,wu2021fairness,wu2021learning}.
For example, Zhang et al.~\shortcite{zhang2018mitigating} proposed to apply adversarial learning to the representations learned by a model by propagating the negative gradients of a sensitive attribute discriminator to the model.
They also proposed to remove the projection of task-specific gradients on the space of discriminator gradients to ensure that optimizing the loss in downstream tasks does not help the discriminator.
Madras et al.~\shortcite{madras2018learning} proposed to use the soft labels predicted by the adversary model and the hidden representations to reconstruct the input data.
Wu et al.~\shortcite{wu2021fairness} proposed a decomposed adversarial learning method that uses a bias-aware user model to capture bias information and uses a bias-free user model to capture bias-independent user interest.
The user embeddings learned by the two models are regularized to be orthogonal.
In these methods, an informative discriminator that can reflect the sensitive attribute space is a necessity for learning fair representations.
However, if only a limited amount of data has sensitive attribute labels, it is difficult to learn an accurate discriminator and thereby the attribute information cannot be effectively removed from hidden representations.
Moreover, existing semi-supervised fair representation learning methods mainly focus on the missing task labels while ignoring the scarcity of data with sensitive attribute labels~\cite{noroozi2019leveraging,zhang2020fairness,brubach2021fairness}.
Different from existing methods, our approach uses a semi-supervised variational autoencoder to exploit useful information from data without attribute labels, which can learn fair representations on limited samples with revealed sensitive attributes.

\subsection{Semi-supervised Variational Autoencoder}

Variational autoencoder~\cite{kingma2013vae} is a widely used generative model developed from the standard autoencoder.
It typically encodes the input data $x$ into a latent space $z$ with Gaussian distributions, and then samples data points from the latent space to construct the input data via a decoder.
Due to the nature of generative models, variational autoencoders can be used for semi-supervised learning, which was first proposed in~\cite{kingma2014semi} (named \textit{Semi-VAE}).
In this model, both the input data $x$ and its label $y$ are used for learning the latent variable $z$ via an encoder $q_\phi(z|x,y)$, where $\phi$ denotes its parameters. 
The decoder then constructs input data using a distribution $p_\theta(x|y,z)$, where $\theta$ is the decoder parameters.
The task label predictor distribution is denoted as $q_\phi(y|x)$.
The distribution of the latent variable $z$ is derived as follows:
\begin{equation}
z\sim q_\phi(z|x,y)=\mathcal{N}(\mu(\mathbf{x},y), diag(\sigma^{2}(\mathbf{x},y))),
\end{equation}
where $\mathbf{x}$ is the hidden representation of $x$ learned by the encoder.

In \textit{Semi-VAE}, on labeled data the evidence lower bound of $x$ with observed label $y$ is  formulated as follows:
\begin{equation}
\begin{aligned}
{\log} p_\theta(x,y)\geq \mathbb{E}_{q_\phi(z|x,y)}[{\log} p_\theta(x|y,z)]+{\log} p_\theta(y)\\-KL(q_\phi(z|x,y)||p(z))=-\mathcal{L}(x,y),
\label{eq01}
\end{aligned}
\end{equation}
where the prior distribution $p(z)$ is typically a standard Gaussian distribution.
On unlabeled data, the label $y$ is given by the classifier, and the evidence lower bound is formulated as follows:
\begin{equation}
\begin{aligned}
{\rm log} p_\theta(x)& \geq \sum_y q_\phi(y|x)(-\mathcal{L}(x,y))+\mathcal{H}(q_\phi(y|x))\\
&= -\mathcal{U}(x),        
\label{eq02}
\end{aligned}
\end{equation}
where $\mathcal{H}$ denotes entropy. 
The unified loss function on the union of labeled and unlabeled data is written as follows:
\begin{eqnarray}
\centering
\mathcal{L}=\sum_{(x,y)\in \mathcal{D}_{l}}{\mathcal{L}(x,y)}+\sum_{x\in \mathcal{D}_{u}}{\mathcal{U}(x)}+ \alpha\mathbb{E}_{(x,y)\in \mathcal{D}_{l}}[-{\rm log} q_\phi(y|x)]
\end{eqnarray}
where $\mathcal{D}_{l}$ and $\mathcal{D}_{u}$ denote the labeled and unlabeled data sets, respectively, and $\alpha$ is a coefficient that controls the relative importance of classification loss. 
By optimizing the loss function of the \textit{Semi-VAE}, the model can be aware of the relatedness between label prediction and input data reconstruction, which can help exploit useful information of unlabeled data to alleviate the scarcity problem of labeled training data.
However,  \textit{Semi-VAE} cannot be directly applied to semi-supervised adversarial learning, because in adversarial learning the discriminator cannot accurately infer the sensitive attributes from hidden representations and thereby cannot help input data reconstruction.
To solve this problem, we propose an adversarial semi-supervised variational autoencoder based on decomposed adversarial learning~\cite{wu2021fairness}, which can mine the relatedness between input data reconstruction and the sensitive attributes predicted by a bias-aware model and meanwhile encourage the bias-free model to be less attribute discriminative.

\section{Semi-supervised Fair Representation Learning}\label{sec:Model}

Next, we introduce the details of our semi-supervised fair representation learning approach named \textit{Semi-FairVAE}.
We will first give a formal definition of the problem studied in this paper, and then introduce the details of our approach.

\begin{table}[!t]
\caption{The main variable denotations in our method.}\label{denotation}
\begin{tabular}{cl}
\Xhline{1.0pt}
\multicolumn{1}{c}{\textbf{Variable}} & \multicolumn{1}{c}{\textbf{Description}}                  \\ \hline
$\mathbf{x}$                 & Input features                                   \\
$\mathbf{\hat{x}}$           & Reconstructed features                           \\
$\mathbf{y}$                 & Real task label                                  \\
$\mathbf{\hat{y}}$           & Predicted task label                             \\
$\mathbf{z}$                 & Real sensitive attribute label                   \\
$\mathbf{\hat{z}}$           & Predicted sensitive attribute label (bias-aware) \\
$\mathbf{\tilde{z}}$          & Predicted sensitive attribute label (bias-free)  \\
$\mathbf{r}_f$               & Bias-free feature representation                 \\
$\mathbf{r}_b$               & Bias-aware feature representation                \\
$\mathbf{r}$                 & Overall feature representation                   \\
$\mathbf{\mu}$               & Mean vector of latent space                      \\
$\mathbf{\sigma}$           & Co-variance vector of latent space               \\
$\mathbf{h}$                 & Latent Representation                            \\
$\mathbf{c}$                 & Combined hidden representation                   \\
$\mathcal{L}_P$                 & Attribute prediction loss               \\
$\mathcal{L}_A$                 & Attribute adversarial loss               \\
$\mathcal{L}_O$                 & Orthogonality loss               \\
$\mathcal{L}_T$                 & Task loss               \\
$\mathcal{L}_R$                 & Reconstruction loss               \\ \Xhline{1.0pt}
\end{tabular}
\end{table}

\subsection{Problem Definition}

In our approach, we denote the input feature of a sample as $\mathbf{x}$ and its label in the target task as $\mathbf{y}$.
The sensitive attribute of this sample is denoted as $\mathbf{z}$ if it is observed.
The entire dataset $\mathcal{D}$ is composed of a set $\mathcal{D}_l$ with observed sensitive attributes and a set $\mathcal{D}_u$ without labeled sensitive attributes.
The goal of the fair representation model is to learn a fair representation $\mathbf{r}_f$ for each sample from $\mathbf{x}$, where its sensitive attribute $\mathbf{z}$ can be minimally inferred from its representation $\mathbf{r}_f$.
In addition, the representations should be maximally informative for predicting the labels in the target task.
We summarize the denotations of variables used in our approach in Table~\ref{denotation}. 
Their details are introduced in the following sections.

\begin{figure*}[!t]
  \centering
    \includegraphics[width=0.85\linewidth]{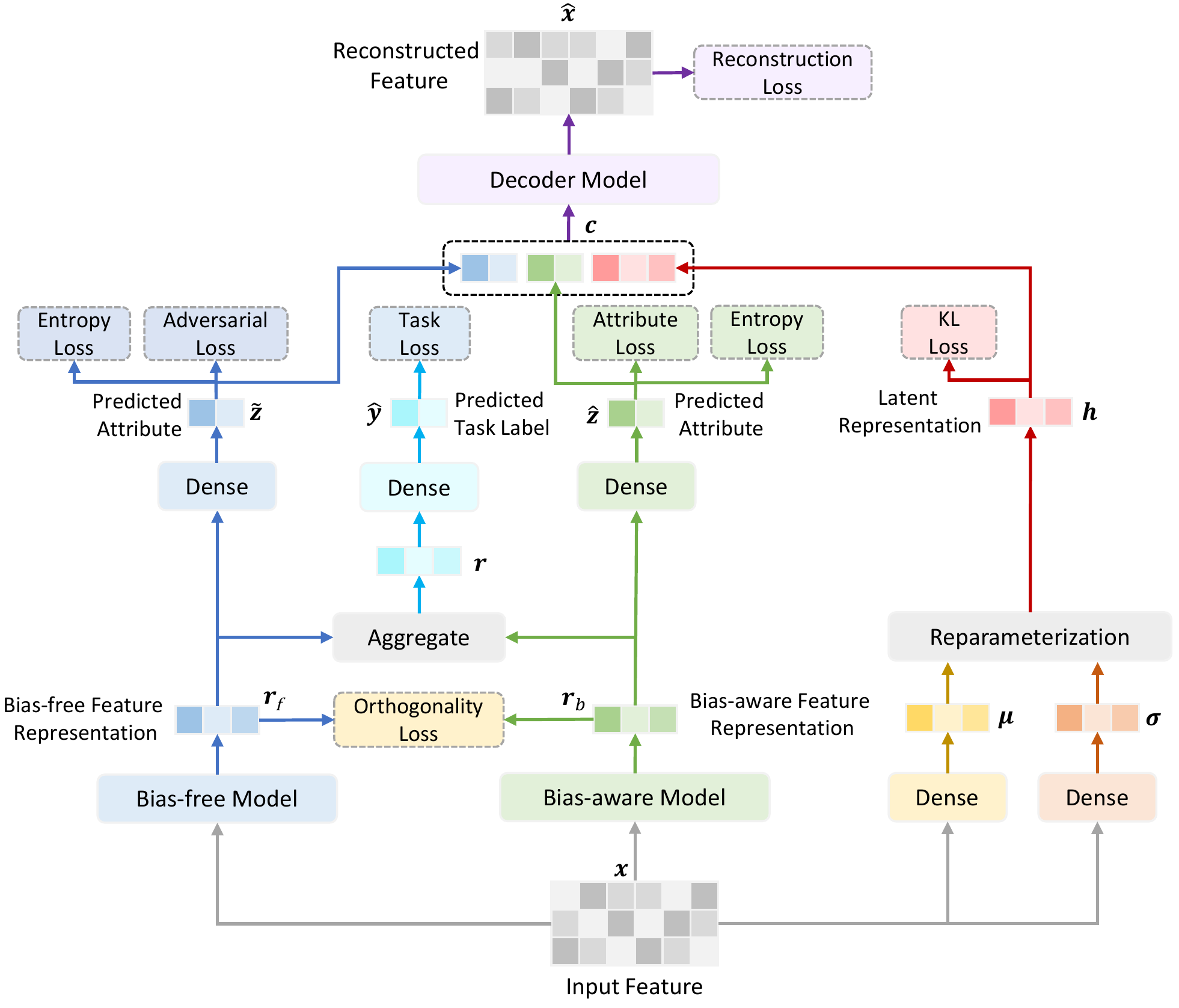}
  \caption{The  framework of our \textit{Semi-FairVAE} approach.}
  \label{fig.model}
\end{figure*}

\subsection{Model Framework}

We then introduce our semi-supervised fair representation learning model.
Its overall framework is shown in Fig.~\ref{fig.model}.
We introduce it in detail in the following paragraphs.

In existing semi-supervised learning frameworks based on VAEs, a prerequisite is learning an informative predictor model based on data with observed labels.
However, in the standard adversarial fair representation learning method, the discriminator is deceived by the feature encoder and it cannot effectively predict the sensitive attributes from feature representations, thereby the vanilla adversarial fair representation learning paradigm is not compatible with semi-supervised VAE.
Motivated by~\cite{wu2021fairness}, we propose to incorporate the decomposed adversarial learning framework into semi-supervised VAE.
As shown in Fig.~\ref{fig.model}, the decomposed adversarial learning framework has a bias-aware model to learn bias-aware feature representations that mainly capture bias information on sensitive attributes, and a bias-free model to learn bias-free representations that encode bias-independent data information.
We denote the bias-free feature representation as $\mathbf{r}_f$ and the bias-aware feature representation as $\mathbf{r}_b$.
To encourage the bias-aware model to maximally capture the bias information on sensitive attributes, an attribute predictor is used to infer sensitive attributes labels $\hat{\mathbf{z}}$ from the hidden representations learned by the bias-aware model.
We apply an attribute prediction loss $\mathcal{L}_P$ to the predictor to optimize its ability in inferring sensitive attributes, which is formulated as:
\begin{equation}
    \mathcal{L}_P=-\sum_i \mathbf{z}_i\log(\hat{\mathbf{z}}_i),
\end{equation}
where $\mathbf{z}_i$ and $\hat{\mathbf{z}}_i$ denote the real and predicted attribute labels for the $i$-th class, respectively.
In addition, to eliminate the bias information encoded by the bias-free model, an adversarial learning module is applied to the hidden representations learned by the bias-free model.
We use a discriminator to infer sensitive attribute labels $\tilde{\mathbf{z}}$ from the bias-free feature representation $\mathbf{r}_f$.
It is used for computing the adversarial loss $\mathcal{L}_A$ for regularizing the model, which is formulated as follows:
\begin{equation}
    \mathcal{L}_A=-\sum_i \mathbf{z}_i\log(\tilde{\mathbf{z}}_i),
\end{equation}
where $\tilde{\mathbf{z}}_i$ is the attribute label for the $i$-th class predicted by the discriminator.
Note that the adversarial loss is used to train the discriminator, while its negative gradients are propagated to the bias-free model.
In order to further purify the bias-free feature representation, an additional orthogonality regularization loss $\mathcal{L}_O$ is applied to the bias-free and bias-aware feature representations to encourage them to be orthogonal with each other, which can be formulated as follows:
\begin{equation}
    \mathcal{L}_O=\frac{|\mathbf{r}_f\cdot\mathbf{r}_b|}{||\mathbf{r}_f||\cdot ||\mathbf{r}_b||}.
\end{equation}
Since the target task may be relevant to the bias information on sensitive attributes, we aggregate the bias-aware  and bias-free feature representations (we use the summation of them for simplicity) into a unified feature representation $\mathbf{r}$.
The task label $\mathbf{\hat{y}}$ is predicted from $\mathbf{r}$ with a task-specific predictor, and the task loss $\mathcal{L}_T$ is formulated as follows:
\begin{equation}
    \mathcal{L}_T=-\sum_i \mathbf{y}_i\log(\hat{\mathbf{y}}_i),
\end{equation}
where $\mathbf{y}_i$ and $\hat{\mathbf{y}}_i$ are the gold and predicted task label for the $i$-th class.
Note that in the test phase, only the bias-free feature representations $\mathbf{r}_f$ are used for predicting the labels of target tasks.
Thus, the method for aggregating the bias-free and bias-aware feature representations should be a linear operation to ensure that the task specific predictor can be used for test.

Following the semi-supervised  variational autoencoder framework, we  use another encoder to encode the input features into the latent space.
We use two dense layers to learn the mean and variance vectors for the input feature, which are computed by:
\begin{equation}
    \mathbf{\mu}=f(\mathbf{W}\mathbf{x}+\mathbf{w}),
\end{equation}
\begin{equation}
    \mathbf{\sigma}=f(\mathbf{U}\mathbf{x}+\mathbf{u}),
\end{equation}
where $\mathbf{W}$, $\mathbf{w}$, $\mathbf{U}$ and $\mathbf{u}$ are parameters, $f(\cdot)$ is the activation function.
The latent representation $\mathbf{h}$ is sampled from the latent space spanned by $\mathbf{\mu}$ and $\mathbf{\sigma}$ via reparameterization as follows:
\begin{equation}
    \mathbf{h}=\mathcal{N}(0, \mathbf{I})*\mathbf{\sigma} + \mathbf{\mu}.
\end{equation}
The latent representation is further concatenated with the sensitive attribute labels predicted by the attribute predictor and the discriminator to form a union $\mathbf{c}=[\tilde{\mathbf{z}}, \hat{\mathbf{z}}, \mathbf{h}]$.
It is used to reconstruct the input data $\mathbf{x}$ through a decoder model.
To train the decoder model to make accurate input data reconstruction, we apply a reconstruction loss\footnote{We apply L2 distance rather than crossentropy because the input features can be continuous variables.} to the reconstructed feature $\mathbf{\hat{x}}$, which is denoted as follows:
\begin{equation}
    \mathcal{L}_R=\frac{1}{d}\sum_i^d(\mathbf{\hat{x}}_i-\mathbf{x}_i)^2,
\end{equation}
where $d$ is the dimension of input feature.
Note that on data with observed sensitive attribute labels, we use the real attribute label $\mathbf{z}$ to replace the predicted labels $\mathbf{\hat{z}}$ in input reconstruction to help capture the joint distribution of sensitive attributes and input data.
In addition, since we expect the sensitive attribute labels predicted by the discriminator to be random, we use soft label vectors with uniform class probability distributions to replace the attribute labels $\mathbf{\hat{z}}$ in reconstructing input data.\footnote{We assume that the expected class distribution is balanced if the attribute labels inferred from hidden representations are random.}
Thus, on data with labeled sensitive attributes, the unified loss function for model training $\mathcal{L}_{labeled}$ can be formulated as follows:
\begin{equation}
    \mathcal{L}_{labeled}(\mathbf{x}, \mathbf{y}, \mathbf{z}) = \mathcal{L}_P- \lambda \mathcal{L}_A+  \mathcal{L}_O+ \mathcal{L}_T + \sum_{(\mathbf{x}, \mathbf{z})\in \mathcal{D}_l}\mathcal{L}(\mathbf{x}, \mathbf{z}),
\end{equation}
\begin{equation}
   \mathcal{L}(\mathbf{x}, \mathbf{z}) = \mathcal{L}_R+ KL(q_\phi(\mathbf{h}|\mathbf{x}, \mathbf{z})||p(\mathbf{h}))-\log_{p_\theta}(\mathbf{z}),
\end{equation}
where $\lambda$ is the adversarial learning strength.
On data without  sensitive attributes, the loss function $\mathcal{L}_{unlabeled}$ is formulated as:
\begin{equation}
    \mathcal{L}_{unlabeled}(\mathbf{x}, \mathbf{y}) = \mathcal{L}_O+ \mathcal{L}_T + \sum_{\mathbf{x}\in \mathcal{D}_u}\mathcal{U}(\mathbf{x}),
\end{equation}
\begin{equation}
   \mathcal{U}(\mathbf{x})  = \sum_\mathbf{z} q_\phi(\mathbf{z}|\mathbf{x})(\mathcal{L}(\mathbf{x},\mathbf{z}))-\mathcal{H}(\mathbf{\hat{z}})-\mathcal{H}(\mathbf{\tilde{z}}).
\end{equation}
In this formula, different from the standard unlabeled loss in Semi-VAE (the first two terms), we further add an additional entropy term of the sensitive attribute labels inferred from the bias-free feature representations.\footnote{Note that we do not regard the union of  $\hat{z}$ and $\tilde{z}$ as a unified latent variable because $\tilde{z}$ is expected to be independent on $\mathbf{x}$ and $\hat{z}$.}
If their entropy is larger, it means that there exists less bias information in the bias-free feature representations and the discriminator can make less confident predictions, which is consistent with the goal of fair representation learning.
By optimizing the loss function $\mathcal{L}_{unlabeled}$ on a large amount of data without sensitive attribute labels, the bias-aware model is encouraged to be more discriminative in inferring sensitive attributes while the bias-free model can be more independent on the bias introduced by sensitive attributes to generate fairer representations.
For model training, we jointly optimize the loss functions on both labeled and unlabeled datasets, which is formulated as follows:
\begin{eqnarray}
\centering
\mathcal{L}=\sum_{(\mathbf{x}, \mathbf{y} ,\mathbf{z})\in \mathcal{D}_{l}}\mathcal{L}_{labeled}(\mathbf{x}, \mathbf{y}, \mathbf{z})+\sum_{(\mathbf{x}, \mathbf{y})\in \mathcal{D}_{u}}\mathcal{L}_{unlabeled}(\mathbf{x}, \mathbf{y}).
\end{eqnarray}
After the model converges, the bias-free feature representations $\mathbf{r}_f$ are used for prediction on the test data.
In this way, the final predictions can be less influenced by the bias information related to sensitive attributes.

\section{Experiments}\label{sec:Experiments}

\subsection{Datasets and Experimental Settings}

\begin{table}[!t]
\caption{Statistics of \textit{Adult} and \textit{NewsRec}.}\label{dataset}
\begin{tabular}{lrlr}
\Xhline{1.0pt}
\multicolumn{4}{c}{\textit{Adult}}                                 \\ \hline
\#train samples & 32,561    & \#test samples  & 16,281    \\ \hline
\multicolumn{4}{c}{\textit{NewsRec}}                               \\ \hline
\#news          & 42,255    & \#users         & 10,000    \\
\#impressions   & 360,428   & avg. title len. & 11.29     \\
\#train logs    & 7,773,027 & \#test logs     & 2,701,466 \\ \Xhline{1.0pt}
\end{tabular}
\end{table}

\begin{table*}[t]
 \caption{Accuracy and fairness of different methods on the \textit{Adult} dataset under different ratios of samples with sensitive attribute labels. Higher accuracy indicates better performance and lower DP or OPP indicates better fairness.} \label{table.performance} 
\begin{tabular}{l|ccc|ccc|ccc}
\Xhline{1.0pt}
\multicolumn{1}{c|}{\multirow{2}{*}{\textbf{Method}}} & \multicolumn{3}{c|}{\textbf{10\%}} & \multicolumn{3}{c|}{\textbf{20\%}} & \multicolumn{3}{c}{\textbf{50\%}} \\ \cline{2-10} 
\multicolumn{1}{c|}{}                        & Acc     & DP     & OPP    & Acc     & DP    & OPP    & Acc     & DP    & OPP    \\ \hline
LR                                           & 0.8484  & 0.1548 & 0.0815 & 0.8484  & 0.1548& 0.0815 &0.8484   & 0.1548& 0.0815  \\
LR+AL                                        & 0.8370  & 0.1428 & 0.0690 & 0.8357  & 0.1343& 0.0654 &0.8364   & 0.1269& 0.0625  \\
LR+AL+ST                                     & 0.8348  & 0.1259 & 0.0616 & 0.8329  & 0.1212& 0.0589 &0.8331   & 0.1174& 0.0588  \\
LR+DAL                                       & 0.8389  & 0.1398 & 0.0668 & 0.8371  & 0.1323& 0.0646 &0.8369   & 0.1247& 0.0619  \\
LR+DAL+ST                                    & 0.8367  & 0.1232 & 0.0609 & 0.8329  & 0.1201& 0.0581 &0.8331   & 0.1158& 0.0576  \\
LR+Semi-FairVAE                               & 0.8366  & 0.1134 & 0.0577 & 0.8344  & 0.1101& 0.0553 &0.8336   & 0.1078& 0.0549  \\ \hline
DNN                                          & 0.8441  & 0.1597 & 0.0876 & 0.8441  & 0.1597& 0.0876 &0.8441   & 0.1597& 0.0876  \\
DNN+AL                                       & 0.8298  & 0.1396 & 0.0688 & 0.8307  & 0.1331& 0.0634 &0.8295   & 0.1241& 0.0606  \\
DNN+AL+ST                                    & 0.8288  & 0.1244 & 0.0603 & 0.8293  & 0.1189& 0.0577 &0.8303   & 0.1165& 0.0570  \\
DNN+DAL                                      & 0.8318  & 0.1352 & 0.0682 & 0.8310  & 0.1289& 0.0631 &0.8322   & 0.1215& 0.0605  \\
DNN+DAL+ST                                   & 0.8307  & 0.1197 & 0.0594 & 0.8303  & 0.1149& 0.0570 &0.8329   & 0.1141&	0.0560  \\
DNN+Semi-FairVAE                              & 0.8323  & 0.1106 & 0.0560 & 0.8305  & 0.1096& 0.0547 &0.8331   & 0.1081& 0.0549  \\ \hline
FM                                           & 0.8508  & 0.1646 & 0.0894 & 0.8508  & 0.1646& 0.0894 &0.8508   & 0.1646& 0.0894  \\
FM+AL                                        & 0.8374  & 0.1455 & 0.0703 & 0.8385  & 0.1360& 0.0681 &0.8362   & 0.1282& 0.0645  \\
FM+AL+ST                                     & 0.8343  & 0.1288 & 0.0632 & 0.8332  & 0.1224& 0.0608 &0.8329   & 0.1181& 0.0592  \\
FM+DAL                                       & 0.8390&0.1409&0.0696&0.8391&0.1325&0.0674&0.8371&0.1258&0.0640  \\
FM+DAL+ST                                    & 0.8359&0.1263&0.0630&0.8357&0.1180&0.0607&0.8352&0.1150&0.0583  \\
FM+Semi-FairVAE                               & 0.8372  & 0.1143 & 0.0584 & 0.8369  & 0.1110& 0.0560 &0.8387   & 0.1090& 0.0551  \\ \Xhline{1.0pt}
\end{tabular}

\end{table*}

We conduct experiments on two datasets for different tasks.
The first dataset is  \textit{Adult}~\cite{kohavi1996scaling}\footnote{https://archive.ics.uci.edu/ml/datasets/adult}, which is a widely used benchmark income prediction dataset for fairness-aware machine learning research.
The task is to infer whether the yearly income of a person is higher than \$50K and gender is regarded as the sensitive attribute.
There are 32,650 male users and 16,192 female users in total.
The input sample in this dataset can be formulated as a feature vector with both categorical and numerical variables.
The second dataset is the news recommendation dataset used in FairRec~\cite{wu2021fairness}, which contains the news click logs of 10,000 users as well as the observed gender labels of a part of users.
Among users with gender labels, there are 2,484 male users and 1,744 female users.
We denote this dataset as \textit{NewsRec}.
Gender is the sensitive attribute in this dataset.
Each sample log in this dataset contains a user's historical clicked news, a candidate news article, and the corresponding click label.
Each impression contains a set of candidate news articles displayed to the same user at a certain time.
The statistics of two datasets are listed in Table~\ref{dataset}.
The training/test sets are randomly divided on the \textit{Adult} dataset, while they are partitioned by time on the \textit{NewsRec} dataset (logs in the last week are used for test).

To simulate the scenario where only a small part of the data has observed sensitive attributes, we reserve different ratios of attribute labels and regard the rest as samples without observed attribute labels.
The hidden dimension of different methods is 256.
Since the input of the \textit{Adult} dataset is a feature vector while is a feature embedding sequence on on \textit{NewsRec}, we use a dense layer as the decoder on \textit{Adult} and use a GRU network as the decoder on \textit{NewsRec}.
Adam~\cite{kingma2014adam} is used for model training and the learning rate is 0.01 on \textit{Adult} and 0.001 on \textit{NewsRec}.
The dropout~\cite{srivastava2014dropout} ratio is 0.2.
On the \textit{Adult} dataset, following prior works~\cite{zhang2020fairness}, we use demographic parity (denoted as DP) and equalized opportunity (denoted as OPP) as the fairness metric and use income prediction accuracy as the performance metric.
On the \textit{NewsRec} dataset, following~\cite{wu2020mind} we use AUC as the performance metric.
In addition, we consider two types of fairness metrics.
The first one is taken from~\cite{wu2021fairness}, which uses the accuracy of gender prediction from the top 5  recommendation results  to measure  fairness.
A higher gender prediction accuracy means worse fairness because the recommendation results are more heavily influenced by sensitive attributes.
The second one is similar to equalized opportunity, which uses the AUC differences between users in different gender groups~\cite{dieterich2016compas} (denoted as $\Delta$-AUC).
A smaller $\Delta$-AUC value means better fairness.
On both datasets, we randomly sample 10\% of training data as validation sets, and tune the hyperparameters of our approach and baselines on them. 
We repeat each experiment 5 times with different random seeds and report the average scores.

\begin{table*}[t]
 \caption{Accuracy and fairness of different methods on the \textit{NewsRec} dataset under different ratios of samples with sensitive attribute labels.  Higher AUC indicates better performance. Lower Acc@5 or higher $\Delta$-AUC indicates better fairness.} \label{table.performance2} 
\begin{tabular}{l|ccc|ccc|ccc}
\Xhline{1.0pt}
\multicolumn{1}{c|}{\multirow{2}{*}{\textbf{Method}}} & \multicolumn{3}{c|}{\textbf{10\%}}                  & \multicolumn{3}{c|}{\textbf{20\%}}                  & \multicolumn{3}{c}{\textbf{50\%}}                   \\ \cline{2-10} 
\multicolumn{1}{c|}{}                                 & AUC             & Acc@5           & $\Delta$-AUC     & AUC             & Acc@5           & $\Delta$-AUC     & AUC             & Acc@5           & $\Delta$-AUC     \\ \hline
NAML                                                  & 0.6220          & 0.6745          & 0.0088          & 0.6220          & 0.6745          & 0.0088          & 0.6220          & 0.6745          & 0.0088          \\
NAML+AL                                               & 0.6159          & 0.6644          & 0.0080          & 0.6152          & 0.6594          & 0.0076          & 0.6126          & 0.6436          & 0.0070          \\
NAML+AL+ST                                            & 0.6149          & 0.6423          & 0.0074          & 0.6139          & 0.6441          & 0.0069          & 0.6131          & 0.6390          & 0.0062          \\
NAML+FairRec                                          & 0.6162          & 0.6196          & 0.0063          & 0.6153          & 0.5889          & 0.0057          & 0.6134          & 0.5475          & 0.0037          \\
NAML+FairRec+ST                                       & 0.6156          & 0.5976          & 0.0048          & 0.6147          & 0.5742          & 0.0039          & 0.6139          & 0.5425          & 0.0026          \\
NAML+Semi-FairVAE                                     & 0.6174          & 0.5634 & 0.0034 & 0.6140          & 0.5423 & 0.0022 & 0.6123          & 0.5290 & 0.0015 \\ \hline
LSTUR                                                 & 0.6279          & 0.6786          & 0.0091          & 0.6279          & 0.6786          & 0.0091          & 0.6279          & 0.6786          & 0.0091          \\
LSTUR+AL                                              & 0.6227          & 0.6678          & 0.0084          & 0.6230          & 0.6626          & 0.0080          & 0.6207          & 0.6483          & 0.0073          \\
LSTUR+AL+ST                                           & 0.6219          & 0.6477          & 0.0075          & 0.6209          & 0.6484          & 0.0072          & 0.6203          & 0.6395          & 0.0064          \\
LSTUR+FairRec                                         & 0.6228          & 0.6204          & 0.0066          & 0.6233          & 0.5911          & 0.0060          & 0.6210          & 0.5498          & 0.0039          \\
LSTUR+FairRec+ST                                      & 0.6225          & 0.5998          & 0.0049          & 0.6217          & 0.5769          & 0.0044          & 0.6204          & 0.5411          & 0.0030          \\
LSTUR+Semi-FairVAE                                    & 0.6217          & 0.5648          & 0.0036          & 0.6196          & 0.5426          & 0.0027          & 0.6189          & 0.5298          & 0.0017          \\ \hline
NRMS                                                  & 0.6287 & 0.6839          & 0.0094          & 0.6287 & 0.6839          & 0.0094          & 0.6287 & 0.6839          & 0.0094          \\
NRMS+AL                                               & 0.6237          & 0.6737          & 0.0088          & 0.6232          & 0.6685          & 0.0079          & 0.6208          & 0.6537          & 0.0075          \\
NRMS+AL+ST                                            & 0.6233          & 0.6517          & 0.0079          & 0.6223          & 0.6544          & 0.0074          & 0.6198          & 0.6412          & 0.0066          \\
NRMS+FairRec                                          & 0.6242          & 0.6248          & 0.0067          & 0.6233          & 0.5931          & 0.0062          & 0.6210          & 0.5537          & 0.0041          \\
NRMS+FairRec+ST                                       & 0.6236          & 0.6031          & 0.0052          & 0.6224          & 0.5787          & 0.0045          & 0.6202          & 0.5414          & 0.0033          \\
NRMS+Semi-FairVAE                                     & 0.6228          & 0.5662          & 0.0035          & 0.6203          & 0.5454          & 0.0029          & 0.6196          & 0.5314          & 0.0019          \\  \Xhline{1.0pt}
\end{tabular}
\end{table*}

\subsection{Performance Evaluation}

We evaluate the performance of different methods in terms of their accuracy and fairness.
On the \textit{Adult} dataset, we compare three widely used  methods for feature based representation learning, including logistic regression (LR), deep neural network (DNN), and factorization machine (FM).\footnote{For logistic regression, we regard the element-wise multiplication between weights and input feature vectors as the hidden representations.}
For the DNN based methods, we use two hidden layers with ReLU activation functions.
On the basis of these methods, we further compare five methods, including: (1)  adversarial learning (denoted as AL), which applies adversarial learning to the representations learned by the model;
(2) adversarial learning with self-training (denoted as AL+ST), which uses an attribute predictor to predict sensitive attributes of samples without attribute labels and add confident predictions to the training set~\cite{noroozi2019leveraging,zhang2020fairness};
(3) decomposed adversarial learning (denoted as DAL)~\cite{wu2021fairness}, which is the basic framework in our method;
(4) decomposed adversarial learning with self-training (denoted as DAL+ST);
(5) Semi-FairVAE, our proposed semi-supervised adversarial learning method.
On the \textit{NewsRec} dataset, we use three widely compared benchmark baselines as basic models, including NAML~\cite{wu2019}, LSTUR~\cite{an2019neural} and NRMS~\cite{wu2019nrms}.
In these methods, candidate news articles in an impression are ranked by their personalized click scores given a target user.
We also compare five methods based on them, including (1) AL, vanilla adversarial learning; (2) AL+ST, adversarial learning with self-training; (3) FairRec~\cite{wu2021fairness}, decomposed adversarial learning for news recommendation; (4) FairRec+ST, combining FairRec with self-training; (5)  Semi-FairVAE, our approach.
We compare model performance and fairness under different amounts of data (i.e., 10\%, 20\% and 50\%) with sensitive attribute labels.
The samples without attribute labels are ignored in adversarial training, but they still participate in task label prediction and input data reconstruction.
The results are shown in Tables~\ref{table.performance} and~\ref{table.performance2}, from which we have the following observations:

First, the methods without fairness awareness usually have better accuracy, while they usually make unfair predictions.
By contrast, fairness-aware methods have some sacrifice on accuracy, while their fairness can be improved.
Second, we find that when the data with observed sensitive attribute labels is scarce, the fairness of purely supervised adversarial learning based methods is  unsatisfactory.
This is because the discriminator cannot be well-tuned and sensitive attribute information cannot be fully removed.
Third, compared to pure supervised methods, semi-supervised methods can consistently achieve better fairness, and the advantage is larger when data with labeled attributes is scarcer.
It shows that mining information from data without labeled sensitive attributes can improve adversarial fair representation learning.
Fourth, decomposed adversarial learning methods can usually achieve better fairness than the vanilla adversarial learning methods.
This is because the main target task may have correlations with sensitive attributes, and it may be suboptimal to apply both adversarial loss and task loss to the same representations.
Finally, our proposed \textit{Semi-FairVAE} approach consistently outperforms self-training in terms of both accuracy and fairness.
Further two-sided t-test shows that the fairness improvement of  \textit{Semi-FairVAE} over other baselines are significant ($p<0.01$).
This is because our approach can enforce the model to better train the sensitive attribute predictor and remove bias information from the bias-free representations by incorporating their predicted attribute labels  into input data reconstruction.
These results verify the effectiveness and generality of our method.

\begin{figure*}[!t]
	\centering
	\includegraphics[width=0.85\linewidth]{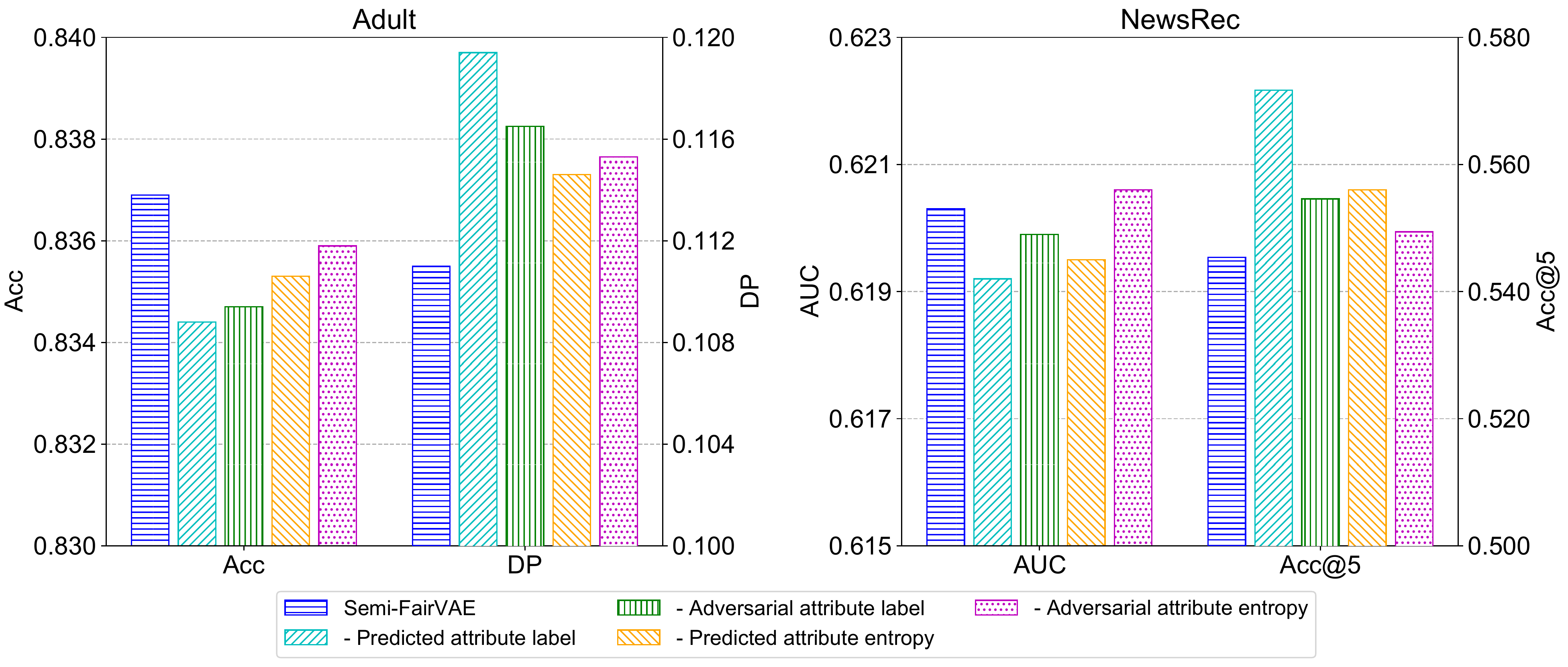}
	
\caption{Influence of removing the attribute labels predicted by bias-aware and bias-free models from input reconstruction and the two associated entropy losses in Eq. (16). Higher Acc or AUC indicates better performance, and lower DP or Acc@5 means better fairness.}\label{fig.ab}
\end{figure*}

\begin{figure*}[!t]
	\centering

	\includegraphics[width=0.85\linewidth]{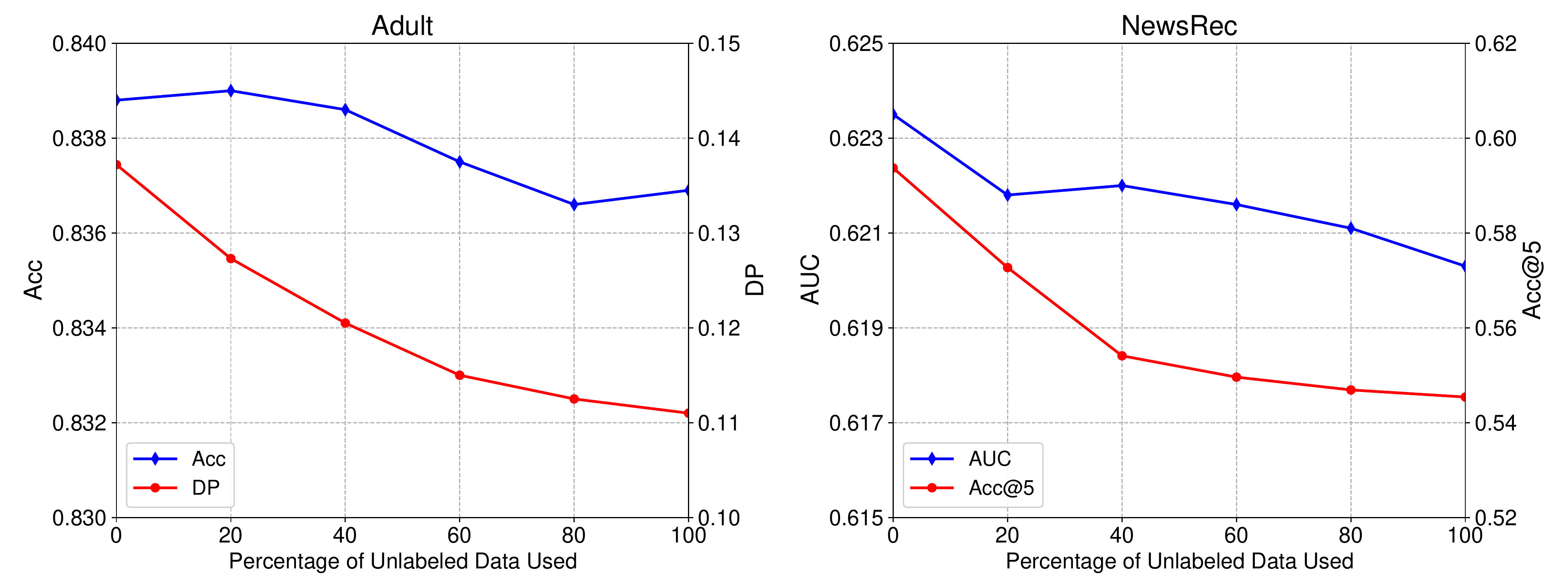}
\caption{Influence of the amount of data without observed sensitive attribute labels on semi-supervised adversarial fair representation learning.}\label{fig.unlabel}
\end{figure*}

\begin{figure*}[!t]
	\centering 
	\includegraphics[width=0.85\linewidth]{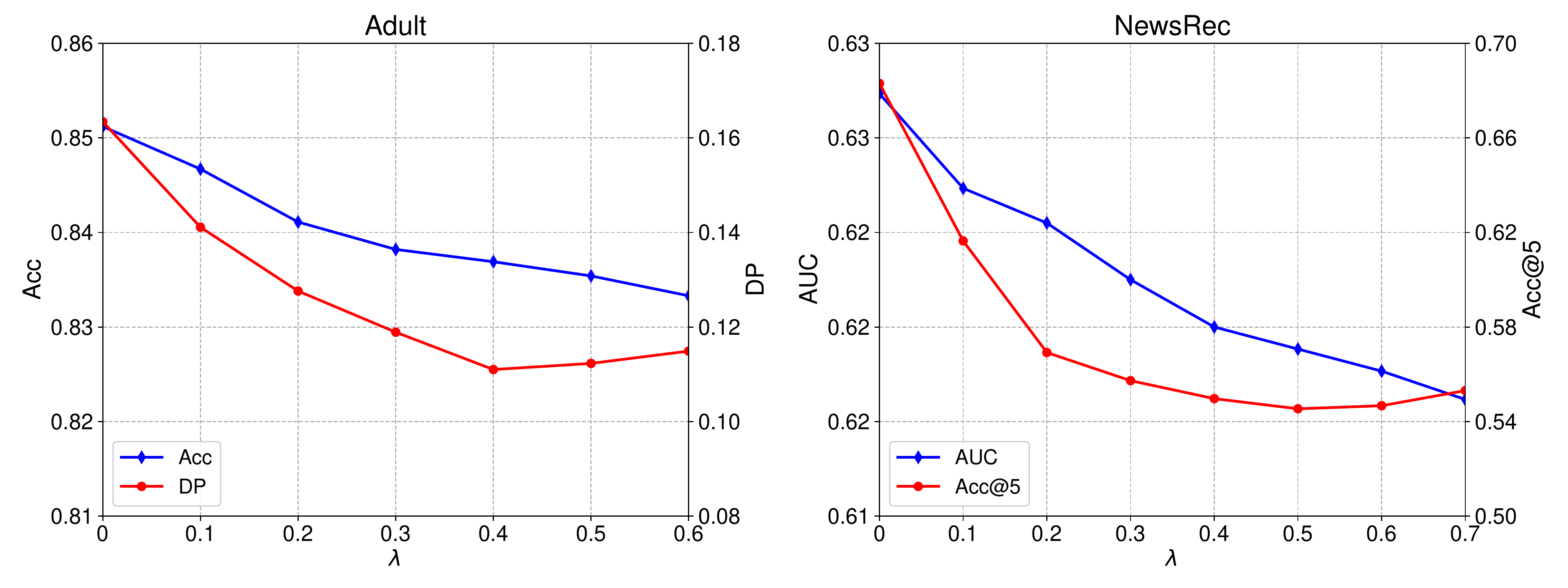}
\caption{Influence of the adversarial loss weight $\lambda$ on accuracy and fairness.}\label{fig.lambda}
\end{figure*}

\begin{figure*}[!t]
	\centering
	\subfigure[W/ semi-supervised learning.]{
	\includegraphics[height=1.9in]{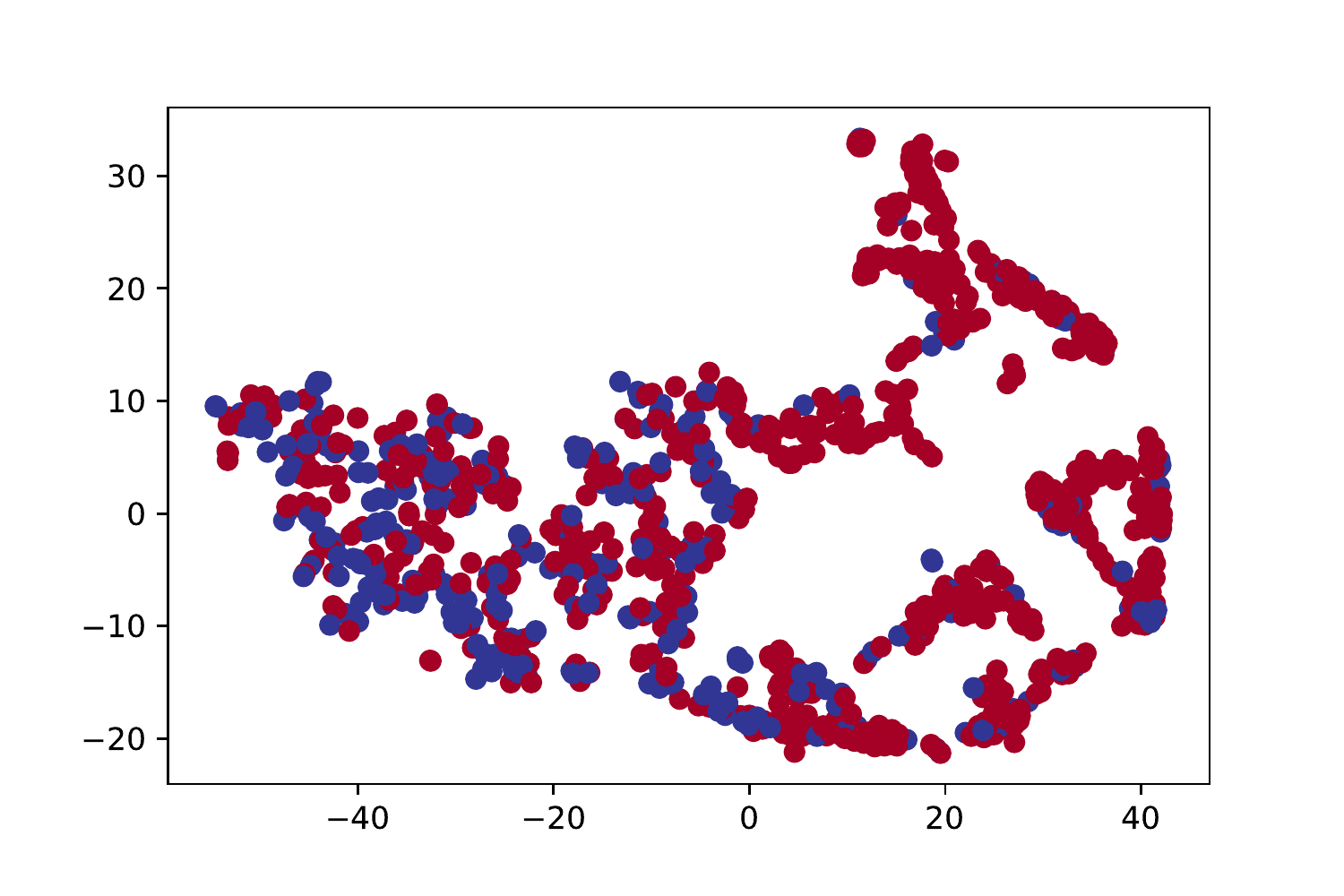}
	}
		\subfigure[W/o semi-supervised learning.]{
	\includegraphics[height=1.9in]{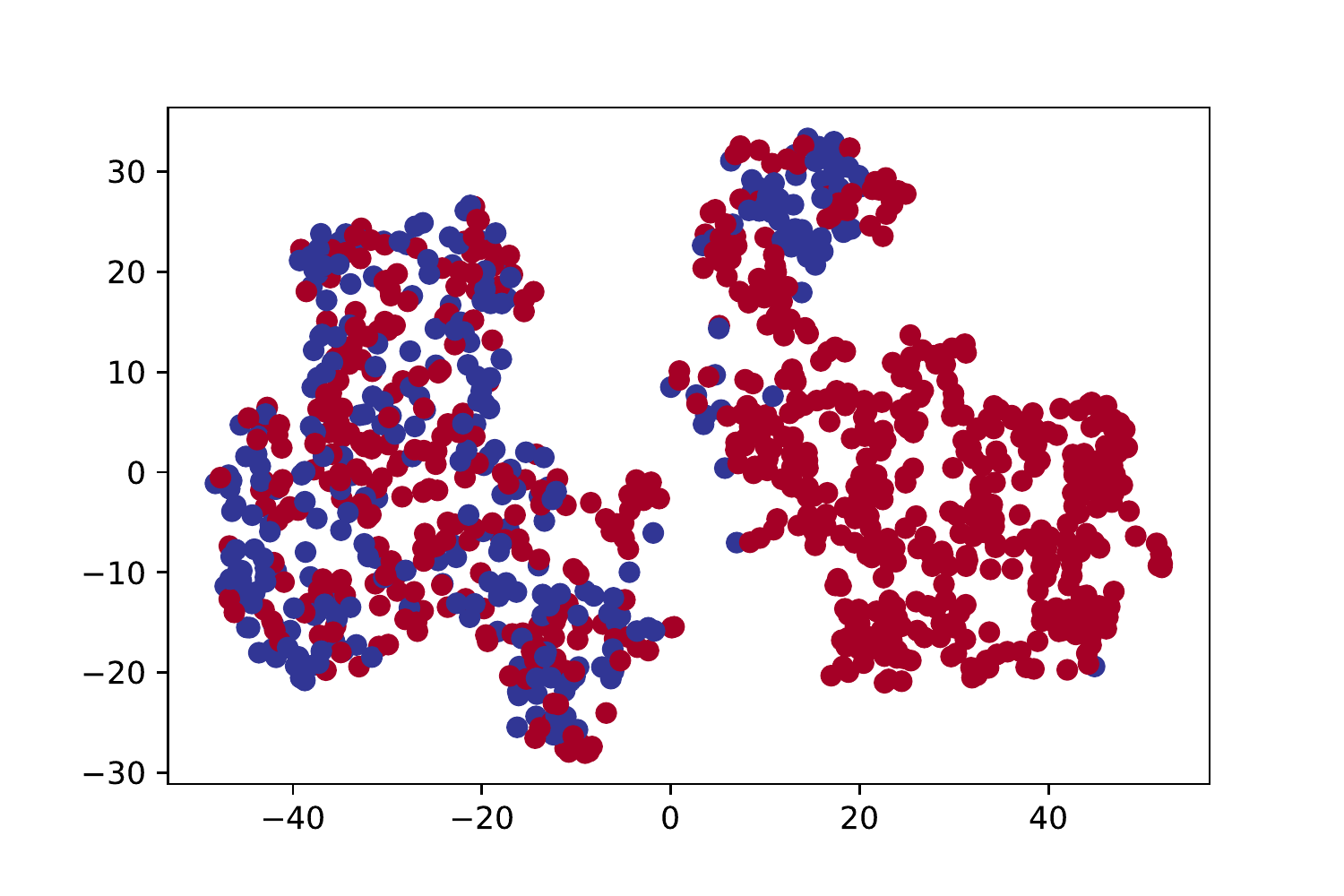}
	}
\caption{Comparison of representations learned by our method and the purely supervised baseline. Red points denote male samples and blue points denote female samples. Better viewed in color.}\label{fig.vis}
\end{figure*}

\subsection{Ablation Study}

Next, we conduct several ablation studies to verify the effectiveness of several key techniques in our approach, including using the attribute labels predicted by the bias-aware and bias-free model for input data reconstruction as well as the entropy losses used for model training on unlabeled data.
We use 20\% of data with attribute labels, and we use FM on \textit{Adult} and NRMS on \textit{NewsRec}.
If not specified, we use these two basic models in the following experiments.
The results on \textit{Adult} and \textit{NewsRec} datasets are shown in Fig.~\ref{fig.ab}.
We have some findings from the results.
First, both the accuracy and fairness decline if the attribute labels inferred from the bias-aware or bias-free feature representations are removed when reconstructing input data.
This shows that reconstructing input data can encourage the model to better capture the bias information on sensitive attributes in the bias-aware model and better remove bias information from the bias-free model.
In addition, if we remove the entropy loss of the attribute labels predicted by the bias-aware model, the accuracy and fairness also drop.
This is because the entropy loss derived from the evidence lower bound of semi-VAE  can encourage the model to make more confident attribute predictions so that the reconstruction can be more accurate, which can help better exploit data without observed attributes.
Besides, removing the negative entropy loss of adversarial attribute label also leads to some fairness degradation.
This may be because this loss can help the bias-free model to better deceive the discriminator and learn bias-independent feature representations.

\subsection{Influence of Unlabeled Data}

Then we study the influence of the amount of  data without sensitive attribute labels on the model accuracy and fairness.
We vary the percentage of unlabeled data used in \textit{Semi-FairVAE} and compare the model performance in Fig.~\ref{fig.unlabel}.
From the results, we find that when more unlabeled data is used, the model fairness can be greatly improved, while the accuracy will slightly decrease.
This is because when more unlabeled data is used, the generative model can better estimate the joint distribution of input features and their sensitive attributes, and thereby can better remove the bias information to improve fairness.
It is also intuitive to have some accuracy sacrifice because removing the bias information also remove some learning shortcuts~\cite{geirhos2020shortcut}.
However, when the amount of unlabeled data is much greater than the number of labeled data, the fairness improvement brought by incorporating more unlabeled data becomes smaller while the performance sacrifice becomes more significant.
This is because the loss on labeled data may not be sufficiently optimized.  
Thus, if used in practical scenarios, we need to choose a proper amount of unlabeled data to participate in the semi-supervised model learning to adjust the tradeoff between accuracy and fairness.

\subsection{Hyperparameter Analysis}

We further analyze the influence of the adversarial loss coefficient $\lambda$ on the model accuracy and fairness. 
We vary the value of $\lambda$ in Eq. (13) and the results are illustrated in Fig.~\ref{fig.lambda}.
We find that when $\lambda$ goes larger, the model accuracy sacrifice becomes larger.
This is because the adversarial loss may affect the model training in the main target task.
In addition, with the increase of $\lambda$, the model fairness first improves and then declines.
This is because if the intensity of adversarial loss is too high, the negative gradients will encode inverse attribute labels into feature representations, which may be harmful to model fairness.
Thus, we choose moderate values for $\lambda$, i.e., 0.4 on \textit{Adult} and 0.5 on \textit{NewsRec}.

\subsection{Representation Visualization}

Finally, we visualize the representations learned by our approach and its variant without semi-supervised learning.
We choose DNN as the basic model and use 20\% of labeled attributes.
The representations learned on \textit{Adult} are shown in Fig.~\ref{fig.vis}.
We can see that the representations learned by purely supervised models severely encode gender biases because many male samples can be discriminated from female samples.
Fortunately, when semi-supervised techniques are incorporated, the representations become less gender-sensitive, which indicates that attribute information encoded in representations is better eliminated.
This result shows the effectiveness of semi-supervised learning in improving fairness-aware representation learning.

\section{Conclusion and Future Work}\label{sec:Conclusion}
In this paper, we propose a semi-supervised fair representation learning method named Semi-FairVAE, which incorporates a semi-supervised adversarial variational autoencoder to reduce the dependency on data with observed sensitive attributes.
In our method, we use a decomposed adversarial learning framework with a bias-aware model to capture sensitive attribute information encoded in input data and a bias-free model to learn fair representations via adversarial learning.
An orthogonal regularization is applied to the hidden representations learned by the two models to better learn fair representations.
In addition, the attribute labels predicted by the two models are further incorporated by a semi-supervised variational autoencoder to reconstruct the input data.
We further regularize the attribute labels predicted from the bias-free model to have high entropy so that  the bias-free model can be more invariant to sensitive attributes.
Extensive experiments on two datasets for different tasks show that our approach can achieve good fairness under scarce data with sensitive attribute labels and meanwhile do not heavily hurt model accuracy.

However, our work also has following limitations.
First, our approach still relies on a certain amount of data with observed sensitive attribute labels.
Thus, it is not compatible with the scenarios where no labeled attribute data is available or there are unseen attribute values in the test data.
Second, our approach may be sensitive to the imbalance of attribute labels. 
If the attribute label is extremely imbalanced, our approach needs to be combined with imbalanced classification techniques.
Third, the attribute label needs to be a categorical variable because the Semi-VAE framework does not support real-valued labels.
Thus, in our future work, we plan to study how to extend our approach to scenarios with few-shot or zero-shot attributes, and explore how to handle the continuous-value sensitive attributes.

\bibliographystyle{ACM-Reference-Format}
\bibliography{main}


\begin{thebibliography}{37}


\ifx \showCODEN    \undefined \def \showCODEN     #1{\unskip}     \fi
\ifx \showDOI      \undefined \def \showDOI       #1{#1}\fi
\ifx \showISBNx    \undefined \def \showISBNx     #1{\unskip}     \fi
\ifx \showISBNxiii \undefined \def \showISBNxiii  #1{\unskip}     \fi
\ifx \showISSN     \undefined \def \showISSN      #1{\unskip}     \fi
\ifx \showLCCN     \undefined \def \showLCCN      #1{\unskip}     \fi
\ifx \shownote     \undefined \def \shownote      #1{#1}          \fi
\ifx \showarticletitle \undefined \def \showarticletitle #1{#1}   \fi
\ifx \showURL      \undefined \def \showURL       {\relax}        \fi
\providecommand\bibfield[2]{#2}
\providecommand\bibinfo[2]{#2}
\providecommand\natexlab[1]{#1}
\providecommand\showeprint[2][]{arXiv:#2}

\bibitem[\protect\citeauthoryear{An, Wu, Wu, Zhang, Liu, and Xie}{An
  et~al\mbox{.}}{2019}]%
        {an2019neural}
\bibfield{author}{\bibinfo{person}{Mingxiao An}, \bibinfo{person}{Fangzhao Wu},
  \bibinfo{person}{Chuhan Wu}, \bibinfo{person}{Kun Zhang},
  \bibinfo{person}{Zheng Liu}, {and} \bibinfo{person}{Xing Xie}.}
  \bibinfo{year}{2019}\natexlab{}.
\newblock \showarticletitle{Neural News Recommendation with Long-and Short-term
  User Representations}. In \bibinfo{booktitle}{\emph{ACL}}.
  \bibinfo{pages}{336--345}.
\newblock


\bibitem[\protect\citeauthoryear{Barocas, Hardt, and Narayanan}{Barocas
  et~al\mbox{.}}{2017}]%
        {barocas2017fairness}
\bibfield{author}{\bibinfo{person}{Solon Barocas}, \bibinfo{person}{Moritz
  Hardt}, {and} \bibinfo{person}{Arvind Narayanan}.}
  \bibinfo{year}{2017}\natexlab{}.
\newblock \showarticletitle{Fairness in machine learning}.
\newblock \bibinfo{journal}{\emph{NIPS tutorial}}  \bibinfo{volume}{1}
  (\bibinfo{year}{2017}), \bibinfo{pages}{2017}.
\newblock


\bibitem[\protect\citeauthoryear{Bengio, Courville, and Vincent}{Bengio
  et~al\mbox{.}}{2013}]%
        {bengio2013representation}
\bibfield{author}{\bibinfo{person}{Yoshua Bengio}, \bibinfo{person}{Aaron
  Courville}, {and} \bibinfo{person}{Pascal Vincent}.}
  \bibinfo{year}{2013}\natexlab{}.
\newblock \showarticletitle{Representation learning: A review and new
  perspectives}.
\newblock \bibinfo{journal}{\emph{TPAMI}} \bibinfo{volume}{35},
  \bibinfo{number}{8} (\bibinfo{year}{2013}), \bibinfo{pages}{1798--1828}.
\newblock


\bibitem[\protect\citeauthoryear{Bengio and LeCun}{Bengio and LeCun}{2015}]%
        {kingma2014adam}
\bibfield{author}{\bibinfo{person}{Yoshua Bengio} {and} \bibinfo{person}{Yann
  LeCun}.} \bibinfo{year}{2015}\natexlab{}.
\newblock \showarticletitle{Adam: {A} Method for Stochastic Optimization}. In
  \bibinfo{booktitle}{\emph{ICLR}}.
\newblock


\bibitem[\protect\citeauthoryear{Brubach, Chakrabarti, Dickerson, Srinivasan,
  and Tsepenekas}{Brubach et~al\mbox{.}}{2021}]%
        {brubach2021fairness}
\bibfield{author}{\bibinfo{person}{Brian Brubach}, \bibinfo{person}{Darshan
  Chakrabarti}, \bibinfo{person}{John~P Dickerson}, \bibinfo{person}{Aravind
  Srinivasan}, {and} \bibinfo{person}{Leonidas Tsepenekas}.}
  \bibinfo{year}{2021}\natexlab{}.
\newblock \showarticletitle{Fairness, Semi-Supervised Learning, and More: A
  General Framework for Clustering with Stochastic Pairwise Constraints}. In
  \bibinfo{booktitle}{\emph{AAAI}}, Vol.~\bibinfo{volume}{35}.
  \bibinfo{pages}{6822--6830}.
\newblock


\bibitem[\protect\citeauthoryear{Calders, Kamiran, and Pechenizkiy}{Calders
  et~al\mbox{.}}{2009}]%
        {calders2009building}
\bibfield{author}{\bibinfo{person}{Toon Calders}, \bibinfo{person}{Faisal
  Kamiran}, {and} \bibinfo{person}{Mykola Pechenizkiy}.}
  \bibinfo{year}{2009}\natexlab{}.
\newblock \showarticletitle{Building classifiers with independency
  constraints}. In \bibinfo{booktitle}{\emph{ICDM Workshops}}. IEEE,
  \bibinfo{pages}{13--18}.
\newblock


\bibitem[\protect\citeauthoryear{Calmon, Wei, Vinzamuri, Ramamurthy, and
  Varshney}{Calmon et~al\mbox{.}}{2017}]%
        {calmon2017optimized}
\bibfield{author}{\bibinfo{person}{Flavio~P Calmon}, \bibinfo{person}{Dennis
  Wei}, \bibinfo{person}{Bhanukiran Vinzamuri},
  \bibinfo{person}{Karthikeyan~Natesan Ramamurthy}, {and}
  \bibinfo{person}{Kush~R Varshney}.} \bibinfo{year}{2017}\natexlab{}.
\newblock \showarticletitle{Optimized pre-processing for discrimination
  prevention}. In \bibinfo{booktitle}{\emph{NIPS}}.
  \bibinfo{pages}{3995--4004}.
\newblock


\bibitem[\protect\citeauthoryear{Dieterich, Mendoza, and Brennan}{Dieterich
  et~al\mbox{.}}{2016}]%
        {dieterich2016compas}
\bibfield{author}{\bibinfo{person}{William Dieterich},
  \bibinfo{person}{Christina Mendoza}, {and} \bibinfo{person}{Tim Brennan}.}
  \bibinfo{year}{2016}\natexlab{}.
\newblock \showarticletitle{COMPAS risk scales: Demonstrating accuracy equity
  and predictive parity}.
\newblock \bibinfo{journal}{\emph{Northpointe Inc}} (\bibinfo{year}{2016}).
\newblock


\bibitem[\protect\citeauthoryear{Donini, Oneto, Ben-David, Shawe-Taylor, and
  Pontil}{Donini et~al\mbox{.}}{2018}]%
        {donini2018empirical}
\bibfield{author}{\bibinfo{person}{Michele Donini}, \bibinfo{person}{Luca
  Oneto}, \bibinfo{person}{Shai Ben-David}, \bibinfo{person}{John
  Shawe-Taylor}, {and} \bibinfo{person}{Massimiliano Pontil}.}
  \bibinfo{year}{2018}\natexlab{}.
\newblock \showarticletitle{Empirical risk minimization under fairness
  constraints}. In \bibinfo{booktitle}{\emph{NIPS}}.
  \bibinfo{pages}{2796--2806}.
\newblock


\bibitem[\protect\citeauthoryear{Filippova}{Filippova}{2012}]%
        {filippova2012user}
\bibfield{author}{\bibinfo{person}{Katja Filippova}.}
  \bibinfo{year}{2012}\natexlab{}.
\newblock \showarticletitle{User demographics and language in an implicit
  social network}. In \bibinfo{booktitle}{\emph{EMNLP}}.
  \bibinfo{pages}{1478--1488}.
\newblock


\bibitem[\protect\citeauthoryear{Geirhos, Jacobsen, Michaelis, Zemel, Brendel,
  Bethge, and Wichmann}{Geirhos et~al\mbox{.}}{2020}]%
        {geirhos2020shortcut}
\bibfield{author}{\bibinfo{person}{Robert Geirhos},
  \bibinfo{person}{J{\"o}rn-Henrik Jacobsen}, \bibinfo{person}{Claudio
  Michaelis}, \bibinfo{person}{Richard Zemel}, \bibinfo{person}{Wieland
  Brendel}, \bibinfo{person}{Matthias Bethge}, {and} \bibinfo{person}{Felix~A
  Wichmann}.} \bibinfo{year}{2020}\natexlab{}.
\newblock \showarticletitle{Shortcut learning in deep neural networks}.
\newblock \bibinfo{journal}{\emph{Nature Machine Intelligence}}
  \bibinfo{volume}{2}, \bibinfo{number}{11} (\bibinfo{year}{2020}),
  \bibinfo{pages}{665--673}.
\newblock


\bibitem[\protect\citeauthoryear{Hu, Zeng, Li, Niu, and Chen}{Hu
  et~al\mbox{.}}{2007}]%
        {hu2007demographic}
\bibfield{author}{\bibinfo{person}{Jian Hu}, \bibinfo{person}{Hua-Jun Zeng},
  \bibinfo{person}{Hua Li}, \bibinfo{person}{Cheng Niu}, {and}
  \bibinfo{person}{Zheng Chen}.} \bibinfo{year}{2007}\natexlab{}.
\newblock \showarticletitle{Demographic prediction based on user's browsing
  behavior}. In \bibinfo{booktitle}{\emph{WWW}}. \bibinfo{pages}{151--160}.
\newblock


\bibitem[\protect\citeauthoryear{Kamiran and Calders}{Kamiran and
  Calders}{2012}]%
        {kamiran2012data}
\bibfield{author}{\bibinfo{person}{Faisal Kamiran} {and} \bibinfo{person}{Toon
  Calders}.} \bibinfo{year}{2012}\natexlab{}.
\newblock \showarticletitle{Data preprocessing techniques for classification
  without discrimination}.
\newblock \bibinfo{journal}{\emph{Knowledge and Information Systems}}
  \bibinfo{volume}{33}, \bibinfo{number}{1} (\bibinfo{year}{2012}),
  \bibinfo{pages}{1--33}.
\newblock


\bibitem[\protect\citeauthoryear{Kamishima, Akaho, Asoh, and Sakuma}{Kamishima
  et~al\mbox{.}}{2012}]%
        {kamishima2012fairness}
\bibfield{author}{\bibinfo{person}{Toshihiro Kamishima},
  \bibinfo{person}{Shotaro Akaho}, \bibinfo{person}{Hideki Asoh}, {and}
  \bibinfo{person}{Jun Sakuma}.} \bibinfo{year}{2012}\natexlab{}.
\newblock \showarticletitle{Fairness-aware classifier with prejudice remover
  regularizer}. In \bibinfo{booktitle}{\emph{ECML-PKDD}}. Springer,
  \bibinfo{pages}{35--50}.
\newblock


\bibitem[\protect\citeauthoryear{Kingma, Mohamed, Rezende, and Welling}{Kingma
  et~al\mbox{.}}{2014}]%
        {kingma2014semi}
\bibfield{author}{\bibinfo{person}{Diederik~P Kingma}, \bibinfo{person}{Shakir
  Mohamed}, \bibinfo{person}{Danilo~Jimenez Rezende}, {and}
  \bibinfo{person}{Max Welling}.} \bibinfo{year}{2014}\natexlab{}.
\newblock \showarticletitle{Semi-supervised learning with deep generative
  models}. In \bibinfo{booktitle}{\emph{NIPS}}. \bibinfo{pages}{3581--3589}.
\newblock


\bibitem[\protect\citeauthoryear{Kingma and Welling}{Kingma and
  Welling}{2014}]%
        {kingma2013vae}
\bibfield{author}{\bibinfo{person}{Diederik~P. Kingma} {and}
  \bibinfo{person}{Max Welling}.} \bibinfo{year}{2014}\natexlab{}.
\newblock \showarticletitle{Auto-Encoding Variational Bayes}. In
  \bibinfo{booktitle}{\emph{ICLR}}.
\newblock


\bibitem[\protect\citeauthoryear{Kohavi et~al\mbox{.}}{Kohavi
  et~al\mbox{.}}{1996}]%
        {kohavi1996scaling}
\bibfield{author}{\bibinfo{person}{Ron Kohavi} {et~al\mbox{.}}}
  \bibinfo{year}{1996}\natexlab{}.
\newblock \showarticletitle{Scaling up the accuracy of naive-bayes classifiers:
  A decision-tree hybrid.}. In \bibinfo{booktitle}{\emph{KDD}},
  Vol.~\bibinfo{volume}{96}. \bibinfo{pages}{202--207}.
\newblock


\bibitem[\protect\citeauthoryear{Lowd and Meek}{Lowd and Meek}{2005}]%
        {lowd2005adversarial}
\bibfield{author}{\bibinfo{person}{Daniel Lowd} {and}
  \bibinfo{person}{Christopher Meek}.} \bibinfo{year}{2005}\natexlab{}.
\newblock \showarticletitle{Adversarial learning}. In
  \bibinfo{booktitle}{\emph{KDD}}. \bibinfo{pages}{641--647}.
\newblock


\bibitem[\protect\citeauthoryear{Madras, Creager, Pitassi, and Zemel}{Madras
  et~al\mbox{.}}{2018}]%
        {madras2018learning}
\bibfield{author}{\bibinfo{person}{David Madras}, \bibinfo{person}{Elliot
  Creager}, \bibinfo{person}{Toniann Pitassi}, {and} \bibinfo{person}{Richard
  Zemel}.} \bibinfo{year}{2018}\natexlab{}.
\newblock \showarticletitle{Learning adversarially fair and transferable
  representations}. In \bibinfo{booktitle}{\emph{ICML}}. PMLR,
  \bibinfo{pages}{3384--3393}.
\newblock


\bibitem[\protect\citeauthoryear{McNamara, Ong, and Williamson}{McNamara
  et~al\mbox{.}}{2019}]%
        {mcnamara2019costs}
\bibfield{author}{\bibinfo{person}{Daniel McNamara},
  \bibinfo{person}{Cheng~Soon Ong}, {and} \bibinfo{person}{Robert~C
  Williamson}.} \bibinfo{year}{2019}\natexlab{}.
\newblock \showarticletitle{Costs and benefits of fair representation
  learning}. In \bibinfo{booktitle}{\emph{AIES}}. \bibinfo{pages}{263--270}.
\newblock


\bibitem[\protect\citeauthoryear{Mehrabi, Morstatter, Saxena, Lerman, and
  Galstyan}{Mehrabi et~al\mbox{.}}{2021}]%
        {mehrabi2021survey}
\bibfield{author}{\bibinfo{person}{Ninareh Mehrabi}, \bibinfo{person}{Fred
  Morstatter}, \bibinfo{person}{Nripsuta Saxena}, \bibinfo{person}{Kristina
  Lerman}, {and} \bibinfo{person}{Aram Galstyan}.}
  \bibinfo{year}{2021}\natexlab{}.
\newblock \showarticletitle{A survey on bias and fairness in machine learning}.
\newblock \bibinfo{journal}{\emph{Comput. Surveys}} \bibinfo{volume}{54},
  \bibinfo{number}{6} (\bibinfo{year}{2021}), \bibinfo{pages}{1--35}.
\newblock


\bibitem[\protect\citeauthoryear{Mikolov, Sutskever, Chen, Corrado, and
  Dean}{Mikolov et~al\mbox{.}}{2013}]%
        {mikolov2013distributed}
\bibfield{author}{\bibinfo{person}{Tomas Mikolov}, \bibinfo{person}{Ilya
  Sutskever}, \bibinfo{person}{Kai Chen}, \bibinfo{person}{Greg~S Corrado},
  {and} \bibinfo{person}{Jeff Dean}.} \bibinfo{year}{2013}\natexlab{}.
\newblock \showarticletitle{Distributed representations of words and phrases
  and their compositionality}. In \bibinfo{booktitle}{\emph{NIPS}}.
  \bibinfo{pages}{3111--3119}.
\newblock


\bibitem[\protect\citeauthoryear{Noroozi, Bahaadini, Sheikhi, Mojab, and
  Philip}{Noroozi et~al\mbox{.}}{2019}]%
        {noroozi2019leveraging}
\bibfield{author}{\bibinfo{person}{Vahid Noroozi}, \bibinfo{person}{Sara
  Bahaadini}, \bibinfo{person}{Samira Sheikhi}, \bibinfo{person}{Nooshin
  Mojab}, {and} \bibinfo{person}{S~Yu Philip}.}
  \bibinfo{year}{2019}\natexlab{}.
\newblock \showarticletitle{Leveraging semi-supervised learning for fairness
  using neural networks}. In \bibinfo{booktitle}{\emph{18th IEEE International
  Conference On Machine Learning And Applications}}. IEEE,
  \bibinfo{pages}{50--55}.
\newblock


\bibitem[\protect\citeauthoryear{Pennington, Socher, and Manning}{Pennington
  et~al\mbox{.}}{2014}]%
        {pennington2014glove}
\bibfield{author}{\bibinfo{person}{Jeffrey Pennington},
  \bibinfo{person}{Richard Socher}, {and} \bibinfo{person}{Christopher~D
  Manning}.} \bibinfo{year}{2014}\natexlab{}.
\newblock \showarticletitle{Glove: Global vectors for word representation}. In
  \bibinfo{booktitle}{\emph{EMNLP}}. \bibinfo{pages}{1532--1543}.
\newblock


\bibitem[\protect\citeauthoryear{Srivastava, Hinton, Krizhevsky, Sutskever, and
  Salakhutdinov}{Srivastava et~al\mbox{.}}{2014}]%
        {srivastava2014dropout}
\bibfield{author}{\bibinfo{person}{Nitish Srivastava},
  \bibinfo{person}{Geoffrey~E Hinton}, \bibinfo{person}{Alex Krizhevsky},
  \bibinfo{person}{Ilya Sutskever}, {and} \bibinfo{person}{Ruslan
  Salakhutdinov}.} \bibinfo{year}{2014}\natexlab{}.
\newblock \showarticletitle{Dropout: a simple way to prevent neural networks
  from overfitting.}
\newblock \bibinfo{journal}{\emph{JMLR}} \bibinfo{volume}{15},
  \bibinfo{number}{1} (\bibinfo{year}{2014}), \bibinfo{pages}{1929--1958}.
\newblock


\bibitem[\protect\citeauthoryear{Van~Engelen and Hoos}{Van~Engelen and
  Hoos}{2020}]%
        {van2020survey}
\bibfield{author}{\bibinfo{person}{Jesper~E Van~Engelen} {and}
  \bibinfo{person}{Holger~H Hoos}.} \bibinfo{year}{2020}\natexlab{}.
\newblock \showarticletitle{A survey on semi-supervised learning}.
\newblock \bibinfo{journal}{\emph{Machine Learning}} \bibinfo{volume}{109},
  \bibinfo{number}{2} (\bibinfo{year}{2020}), \bibinfo{pages}{373--440}.
\newblock


\bibitem[\protect\citeauthoryear{Wu, Wu, An, Huang, Huang, and Xie}{Wu
  et~al\mbox{.}}{2019a}]%
        {wu2019}
\bibfield{author}{\bibinfo{person}{Chuhan Wu}, \bibinfo{person}{Fangzhao Wu},
  \bibinfo{person}{Mingxiao An}, \bibinfo{person}{Jianqiang Huang},
  \bibinfo{person}{Yongfeng Huang}, {and} \bibinfo{person}{Xing Xie}.}
  \bibinfo{year}{2019}\natexlab{a}.
\newblock \showarticletitle{Neural News Recommendation with Attentive
  Multi-View Learning}. In \bibinfo{booktitle}{\emph{IJCAI}}.
  \bibinfo{pages}{3863--3869}.
\newblock


\bibitem[\protect\citeauthoryear{Wu, Wu, Ge, Qi, Huang, and Xie}{Wu
  et~al\mbox{.}}{2019b}]%
        {wu2019nrms}
\bibfield{author}{\bibinfo{person}{Chuhan Wu}, \bibinfo{person}{Fangzhao Wu},
  \bibinfo{person}{Suyu Ge}, \bibinfo{person}{Tao Qi},
  \bibinfo{person}{Yongfeng Huang}, {and} \bibinfo{person}{Xing Xie}.}
  \bibinfo{year}{2019}\natexlab{b}.
\newblock \showarticletitle{Neural News Recommendation with Multi-Head
  Self-Attention}. In \bibinfo{booktitle}{\emph{EMNLP}}.
  \bibinfo{pages}{6390--6395}.
\newblock


\bibitem[\protect\citeauthoryear{Wu, Wu, Wang, Huang, and Xie}{Wu
  et~al\mbox{.}}{2021b}]%
        {wu2021fairness}
\bibfield{author}{\bibinfo{person}{Chuhan Wu}, \bibinfo{person}{Fangzhao Wu},
  \bibinfo{person}{Xiting Wang}, \bibinfo{person}{Yongfeng Huang}, {and}
  \bibinfo{person}{Xing Xie}.} \bibinfo{year}{2021}\natexlab{b}.
\newblock \showarticletitle{FairRec: Fairness-aware News Recommendation with
  Decomposed Adversarial Learning}. In \bibinfo{booktitle}{\emph{AAAI}},
  Vol.~\bibinfo{volume}{35}. \bibinfo{pages}{4462--4469}.
\newblock


\bibitem[\protect\citeauthoryear{Wu, Qiao, Chen, Wu, Qi, Lian, Liu, Xie, Gao,
  Wu, et~al\mbox{.}}{Wu et~al\mbox{.}}{2020}]%
        {wu2020mind}
\bibfield{author}{\bibinfo{person}{Fangzhao Wu}, \bibinfo{person}{Ying Qiao},
  \bibinfo{person}{Jiun-Hung Chen}, \bibinfo{person}{Chuhan Wu},
  \bibinfo{person}{Tao Qi}, \bibinfo{person}{Jianxun Lian},
  \bibinfo{person}{Danyang Liu}, \bibinfo{person}{Xing Xie},
  \bibinfo{person}{Jianfeng Gao}, \bibinfo{person}{Winnie Wu}, {et~al\mbox{.}}}
  \bibinfo{year}{2020}\natexlab{}.
\newblock \showarticletitle{MIND: A Large-scale Dataset for News
  Recommendation}. In \bibinfo{booktitle}{\emph{ACL}}.
  \bibinfo{pages}{3597--3606}.
\newblock


\bibitem[\protect\citeauthoryear{Wu, Chen, Shao, Hong, Wang, and Wang}{Wu
  et~al\mbox{.}}{2021a}]%
        {wu2021learning}
\bibfield{author}{\bibinfo{person}{Le Wu}, \bibinfo{person}{Lei Chen},
  \bibinfo{person}{Pengyang Shao}, \bibinfo{person}{Richang Hong},
  \bibinfo{person}{Xiting Wang}, {and} \bibinfo{person}{Meng Wang}.}
  \bibinfo{year}{2021}\natexlab{a}.
\newblock \showarticletitle{Learning Fair Representations for Recommendation: A
  Graph-based Perspective}. In \bibinfo{booktitle}{\emph{WWW}}.
  \bibinfo{pages}{2198--2208}.
\newblock


\bibitem[\protect\citeauthoryear{Yao and Huang}{Yao and Huang}{2017}]%
        {yao2017beyond}
\bibfield{author}{\bibinfo{person}{Sirui Yao} {and} \bibinfo{person}{Bert
  Huang}.} \bibinfo{year}{2017}\natexlab{}.
\newblock \showarticletitle{Beyond parity: Fairness objectives for
  collaborative filtering}. In \bibinfo{booktitle}{\emph{NIPS}}.
  \bibinfo{pages}{2921--2930}.
\newblock


\bibitem[\protect\citeauthoryear{Zafar, Valera, Gomez~Rodriguez, and
  Gummadi}{Zafar et~al\mbox{.}}{2017}]%
        {zafar2017fairness}
\bibfield{author}{\bibinfo{person}{Muhammad~Bilal Zafar},
  \bibinfo{person}{Isabel Valera}, \bibinfo{person}{Manuel Gomez~Rodriguez},
  {and} \bibinfo{person}{Krishna~P Gummadi}.} \bibinfo{year}{2017}\natexlab{}.
\newblock \showarticletitle{Fairness beyond disparate treatment \& disparate
  impact: Learning classification without disparate mistreatment}. In
  \bibinfo{booktitle}{\emph{WWW}}. \bibinfo{pages}{1171--1180}.
\newblock


\bibitem[\protect\citeauthoryear{Zemel, Wu, Swersky, Pitassi, and Dwork}{Zemel
  et~al\mbox{.}}{2013}]%
        {zemel2013learning}
\bibfield{author}{\bibinfo{person}{Rich Zemel}, \bibinfo{person}{Yu Wu},
  \bibinfo{person}{Kevin Swersky}, \bibinfo{person}{Toni Pitassi}, {and}
  \bibinfo{person}{Cynthia Dwork}.} \bibinfo{year}{2013}\natexlab{}.
\newblock \showarticletitle{Learning fair representations}. In
  \bibinfo{booktitle}{\emph{ICML}}. PMLR, \bibinfo{pages}{325--333}.
\newblock


\bibitem[\protect\citeauthoryear{Zhang, Lemoine, and Mitchell}{Zhang
  et~al\mbox{.}}{2018}]%
        {zhang2018mitigating}
\bibfield{author}{\bibinfo{person}{Brian~Hu Zhang}, \bibinfo{person}{Blake
  Lemoine}, {and} \bibinfo{person}{Margaret Mitchell}.}
  \bibinfo{year}{2018}\natexlab{}.
\newblock \showarticletitle{Mitigating unwanted biases with adversarial
  learning}. In \bibinfo{booktitle}{\emph{AIES}}. \bibinfo{pages}{335--340}.
\newblock


\bibitem[\protect\citeauthoryear{Zhang, Li, Han, Zhou, Yu, et~al\mbox{.}}{Zhang
  et~al\mbox{.}}{2020}]%
        {zhang2020fairness}
\bibfield{author}{\bibinfo{person}{Tao Zhang}, \bibinfo{person}{Jing Li},
  \bibinfo{person}{Mengde Han}, \bibinfo{person}{Wanlei Zhou},
  \bibinfo{person}{Philip Yu}, {et~al\mbox{.}}}
  \bibinfo{year}{2020}\natexlab{}.
\newblock \showarticletitle{Fairness in semi-supervised learning: Unlabeled
  data help to reduce discrimination}.
\newblock \bibinfo{journal}{\emph{TKDE}} (\bibinfo{year}{2020}).
\newblock


\bibitem[\protect\citeauthoryear{Zhao, Zhou, Li, Wang, and Chang}{Zhao
  et~al\mbox{.}}{2018}]%
        {zhao2018learning}
\bibfield{author}{\bibinfo{person}{Jieyu Zhao}, \bibinfo{person}{Yichao Zhou},
  \bibinfo{person}{Zeyu Li}, \bibinfo{person}{Wei Wang}, {and}
  \bibinfo{person}{Kai-Wei Chang}.} \bibinfo{year}{2018}\natexlab{}.
\newblock \showarticletitle{Learning Gender-Neutral Word Embeddings}. In
  \bibinfo{booktitle}{\emph{EMNLP}}. \bibinfo{pages}{4847--4853}.
\newblock


\end{thebibliography}

\end{document}